\documentclass[runningheads]{llncs}
\usepackage{graphicx}
\usepackage{amssymb}
\usepackage{booktabs} 
\usepackage{bm}
\usepackage{multirow}
\usepackage{xspace}
\usepackage[tight,footnotesize]{subfigure}
\usepackage{algorithm}   
\usepackage{amsmath}
\usepackage{enumitem}
\usepackage{algorithmic}
\usepackage{balance}
\usepackage{cancel}
\newcommand{\paratitle}[1]{\vspace{1.5ex}\noindent\textbf{#1}}
\newcommand{\ie}{\emph{i.e.,}\xspace}

\newcommand{\eg}{\emph{e.g.,}\xspace}

\newcommand{\wrt}{\emph{w.r.t.}\xspace}

\newcommand{\model}{HIER}
\begin{document}
\title{Your decision path does matter in pre-training industrial recommenders with multi-source behaviors}


\author{
Chunjing Gan\orcidID{0009-0006-7926-6320}\inst{\star} \and
Binbin Hu\orcidID{0000-0002-2505-1619}\thanks{Equal contribution.} \and
Bo Huang\orcidID{0000-0001-8875-0639} \and
Ziqi Liu\orcidID{0000-0002-4112-3504} \and
Jian Ma\orcidID{0009-0003-1065-506X} \and
Zhiqiang Zhang\orcidID{0000-0002-2321-7259} \and
Wenliang Zhong\orcidID{0009-0006-8861-9503} \and
Jun Zhou\orcidID{0000-0001-6033-6102} 
}
%
%
\institute{Ant Group
\email{\{cuibing.gcj,bin.hbb,yunpo.hb,ziqiliu,mj.mj,lingyao.zzq,yice.zwl, \\jun.zhoujun\}@antgroup.com}}

%
\maketitle              
\begin{abstract}
Online service platforms offering a wide range of services through miniapps have become crucial for users who visit these platforms with clear intentions to find services they are interested in. Aiming at effective content delivery, cross-domain recommendation are introduced to learn high-quality representations by transferring behaviors from data-rich scenarios.
However, these methods overlook the impact of the decision path that users take when conduct behaviors, that is, users ultimately exhibit different behaviors based on various intents.
To this end, we propose \textbf{\model}, a novel \underline{H}ierarchical dec\underline{I}sion path \underline{E}nhanced \underline{R}epresentation learning for cross-domain recommendation.
With the help of graph neural networks for high-order topological information of the knowledge graph between multi-source behaviors, we further adaptively learn
decision paths through well-designed exemplar-level and information bottleneck based contrastive learning.
Extensive experiments in online and offline environments show the superiority of {\model}.

\end{abstract}

\section{Introduction}
Recently integrated online service platforms such as Alipay, WeChat and Tiktok have spread across every corner of life and are becoming more and more important. 
On the whole, online service platforms can bring convenience for both service providers and customers to a great extent, where making appropriate recommendations are at its heart for user experience improvement.
However, making recommendations so as to satisfy the efficient content delivery of both sides still encounters quite a lot difficulties, especially for newly launched domain-agnostic scenarios since it is difficult or even impossible to train a recommender from scratch when there is a small amount of interaction records or even no interaction available.

\begin{figure}[h]
\vspace{-2em}
    \centering
    \includegraphics[width=0.7\columnwidth]{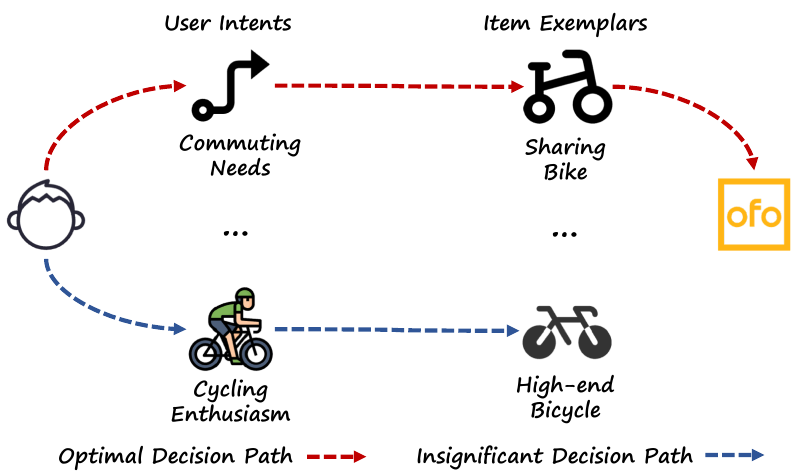}
    \caption{A toy example of possible user decision path when interacts with a sharing-bike.}
    \label{fig:eg}
    \vspace{-2em}
\end{figure}

One possible and plausible recipe is cross-domain recommendation (CDR) \cite{cdrsurvey2021,cdrsurvey2022}, which utilize the behaviors of data-richer source domains (\eg global behaviors that accumulate in online service platforms) for learning transferable knowledge to help improve recommendation in data-sparser target domains (\eg miniapps). 
Existing CDR methods often assume that the availability of both source and target domains for learning mapping function across domains given the shared information ~\cite{lu2023three,commonfeatures2020} or for joint optimization ~\cite{li2023one,zhao2022multi}. 
However, in real-world service platforms, usually source and target domains can be highly diversified and sometimes the target domains are even not available during training, which influences the validity of this assumption and further affects the application of existing CDR approaches. 
Recently, several approaches attempt to incorporate auxiliary information\cite{tiger2022} such as knowledge graph to be the bridge among diversified domains and introduce pre-training~\cite{peterrec2020} into cross-domain recommendation. Specifically, ~\cite{gao2023leveraging,tiger2022,liu2023pre} learn universal user representations via interaction records across multiple source domains and bridging them through auxiliary information such as knowledge graph, text and so on, hoping that the derived representation can be applied to various downstream target scenarios directly or via fine-tuning. 
Despite the considerable performance improvement they have achieved in relatively homogeneous cases such as social and e-commerce platforms, they overlook the impact of the decision path that users take when conduct behaviors, that is, users ultimately exhibit different behaviors based on various intents (A toy example is illustrated in Figure \ref{fig:eg}, when a person with commuting needs arrives, he / she tend to choose a sharing-bike. However, when the intent of cycling enthusiasm dominates, he / she tend to  choose a high-end bicycle.), especially in online service platforms integrated with heterogeneous services / miniapps. 

Given the fore-mentioned limitations of current approaches, 
we seek to take user decision path into consideration when pre-training industrial recommenders with multi-source behaviors and applies to downstream domain-agnostic scenarios.
However, the solution is quite non-trivial, which needs to tackle the following essential challenges in real-world service platforms: 
i) In online service platforms, usually information are unshared among diversified miniapps launched by various service providers, not to mention related items without identical properties. Therefore, there is necessity to narrow down the domain gap and introduce auxiliary information for the bridging of multiple data sources so that the relationships of users, items can be captured in a better way.
ii) Despite the information gain that the bridging process provides, simply aggregating these information indistinguishably may limit its application in a variety of target domains. In fact, the decision path of users is often overlooked. Hence, it is necessary to explicitly take it into consideration thus better capturing user interaction patterns.

To this end, we propose {\model} that takes user decision path into consideration when pre-training industrial recommenders with multi-source
behaviors and applies to downstream domain-agnostic scenarios.
To begin with, we built {\model} upon a knowledge graph bridging users/items  across different domains and then utilize graph neural networks to obtain initial node representations enhanced by the topological structure in a coarse-grained manner.
Towards fine-grained modelling, we define the 
decision path network from
user intents to item exemplars and optimize the decision path selection
via sparsity regularization, in the meantime we optimize the node representations at both ends of the decision paths via exemplar-level and information bottleneck
based contrastive learning.
In the fine-tuning stage, after effortlessly loading pre-trained parameters, {\model} represents users with the universal encoding and demystifies target items in target domains with associated entities, followed by carefully fine-tuning for few-shot and strictly zero-shot recommendation in a lightweight manner.
Comprehensive experiments in both offline and online environments demonstrate that {\model} significantly surpasses a series of state-of-the-art approaches for cross-domain recommendation in both few-shot and zero-shot setting.

\section{The proposed approach}

\begin{figure}[t]
\vspace{-1em}
    \centering
    \includegraphics[width=0.95\columnwidth]{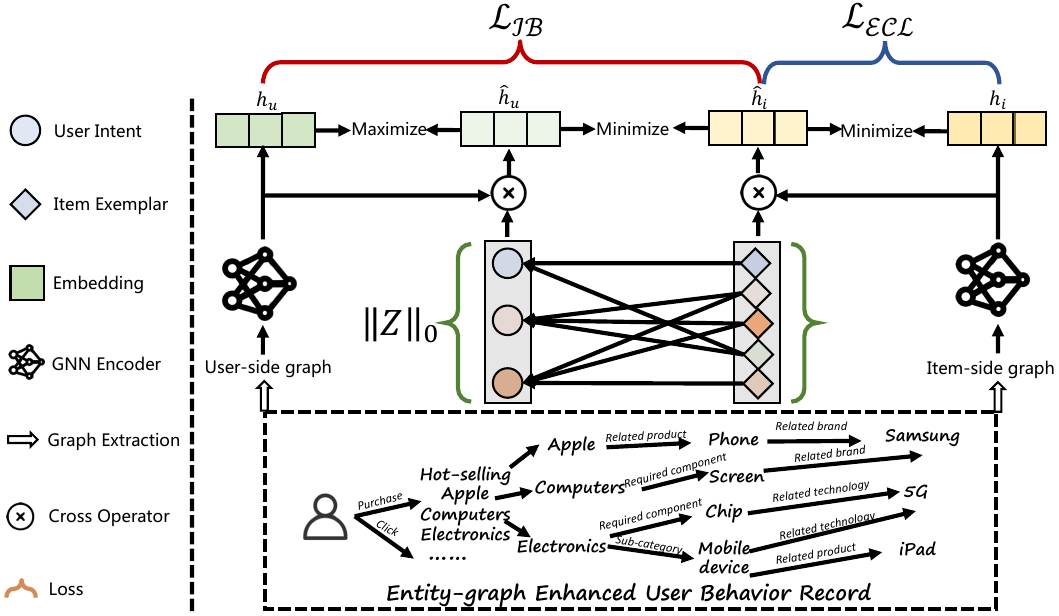}
    \caption{Overall architecture of {\model}.}
    \label{fig:model}
    \vspace{-1em}
\end{figure}
\subsection{Overview}
General Graph Neural Network (GNN) based recommender systems \cite{lightgcn2020,wu2022graph} adopt the classical message-passing scheme: update the center node by aggregating neighbor nodes $\rightarrow$ refine node representations by pooling node representations across multiple layers and then the model is trained according to downstream tasks such as CTR optimization or multi-task optimization.

Inspired by the huge success of GNN-based recommender systems, in this paper, we built {\model} upon a knowledge graph that helps to better capture / enhance the relevance of users, items\footnote{In the pre-training stage, all items are demystified to entities of atom form so that agnostic items in target domains can be represented through entities.}, entities via the high-order graph structure information of the knowledge graph in a coarse-grained manner. Specifically, for a given node $t$ of a specific type  (can be either a user node or an item node), it first aggregates the neighboring entity nodes (Eq.\ref{eq:pretrain_ht}) and enhances the representations of entity nodes by introducing their related entities in the knowledge graph (Eq.\ref{eq:pretrain_he}). Finally, after $L$\footnote{Due to the latency requirements of recommender systems in industrial scenarios, we restrict the number of graph convolution layers to 2, \ie ``user $\rightarrow$ entity $\rightarrow$ entity'' for user nodes and ``entity $\rightarrow$ entity $\rightarrow$ entity'' for item node in our case.} layers of graph convolution, we can combine the embeddings of each layer to obtain the final embedding of node $t$ via Eq.\ref{eq:pretrain_hue_final}. 

\begin{equation}
        {h}_{t}^{(l+1)} = \text{AGG}\left ( {h}_{t}^{(l)},\left \{ {h}_{e}^{(l)}:e \in \mathcal{N}_{t}^{t-e} \right \}\right ).
        \label{eq:pretrain_ht}
\end{equation}

\begin{equation}
        {h}_{e}^{(l+1)} = \text{AGG}\left ( {h}_{e}^{(l)},\left \{ {h}_{e^{'}}^{(l)}:e^{'} \in \mathcal{N}_{e}^{e-e} \right \}\right ).
        \label{eq:pretrain_he}
\end{equation}

\begin{equation}
        \bm{h}_{t} =  \sum_{l=0}^{L} \alpha_l h_{t}^{l}.
        \label{eq:pretrain_hue_final}
\end{equation}

Here, ${h}_{e}^{(0)}$ is a 32-dimensional embedding from a pre-trained BERT, the ``AGG" operation is implemented as a simple weighted sum aggregator and it can be easily implemented as an LSTM aggregator or a bilinear interaction aggregator, and the $\alpha_l$ parameter denotes the significance of layer $l$ \wrt the final representation. 
After obtaining the representations of users and items
respectively, we can easily obtain the preference score of a user towards an item and the cross-entropy based optimization as follows:

\begin{equation}
        \hat{y}_{ui} = \sigma(\bm{h}_{u}^{T} \bm{h}_i).
        \label{eq:pretrain_yue}
\end{equation}

\begin{equation}
        \mathcal{L}_{\mathcal{ET}} = \sum_{\mathcal{S} \in \mathbb{S}}\sum_{\langle u, i, y_{ui}\rangle  \in \hat{\mathcal{H}}^{\mathcal{S}}} \mathcal{C}(y_{ui}, \hat{y}_{ui}).
        \label{eq:loss_et}
\end{equation}

Obviously, in order to predict $\hat{y}_{ui}$ more accurately, the key factors are the two-side representations $\bm{h}_{u}$ and $\bm{h}_i$, which we aim to optimize in Section~\ref{subsec:model_opt_ue} via hierarchical decision path. After that, we finally propose the learning framework of {\model} in Section~\ref{subsec:model_loss}.

\subsection{Hierarchical Decision Path Enhanced Representation Learning}\label{subsec:model_opt_ue}

\subsubsection{Improving Item Representation via Exemplar-level Contrastive Learning}\label{subsec:model_opt_e}
Leveraging entity-centric contexts has proven effective for transferable recommendations, yet there is room to enhance universal knowledge acquisition by clustering entities sharing traits (\eg ``Adidas'' and ``Nike'') and distancing those that differ significantly (\eg ``Adidas'' and ``Coffee'').
Drawing inspiration from cluster analysis, the integration of exemplar-based concepts into pre-training for recommenders aims to group similar entities in the representation space, which not only enriches the recommender with cluster-level semantics for better generalization but also strengthens its resilience against the unavoidable noise present in real-world settings.

To this end, we propose a contrastive exemplar learning module to improve entity representation. 
Specifically, we adopt a $d$-dimensional learnable matrix $\mathbf{P} \in \mathbb{R}^{n \times d}$ 
to denote the latent exemplars in the knowledge graph, where $n$ is the number of exemplars. 
For each item $i$, we calculate the normalized similarity vector \wrt the $n$ exemplars as follows:
\begin{equation}
    \bm{s}_i = \text{SoftMax}(\mathbf{P} \cdot \bm{h}_i).
    \label{eq:sim}
\end{equation}

For an effective learning process utilizing exemplars, our objective is that the chosen exemplar can assist in ``predicting'' the target item through a weighted sum based on similarity scores. Therefore, given the target item $i$, we derive its exemplar-based perspective as follows:
\begin{equation}
\label{eq:pt_view}
    \hat{\bm{h}}_i = \mathbf{P}^T \cdot \bm{s}_i.
\end{equation}

After that, the goal of the proposed contrastive exemplar learning is to pull related views close and push others away, we define the loss based on InfoNCE
as follows:
\begin{equation}
 \label{eq:loss_pcl}
    \begin{split}
    \mathcal{L}_{ECL} &= -\sum_{i \in \mathcal{I}} \left(  \log{\frac{\exp{(\delta(\bm{h}_i, \hat{\bm{h}}_i)/\tau)}}{\sum_{i' \sim \mathcal{P}_{neg}}\exp{(\delta(\bm{h}_i, \hat{\bm{h}}_{i'})/\tau)}}} \right. \\
    & \left. + \log{\frac{\exp{(\delta(\hat{\bm{h}}_i, \bm{h}_i)/\tau)}}{\sum_{i' \sim \mathcal{P}_{neg}}\exp{(\delta( \hat{\bm{h}}_i, \bm{h}_{i'})/\tau)}}}  \right), \\
    \end{split}
\end{equation}
where $\mathcal{P}_{neg}$ is the noise distribution for generating negative views, which could be set as uniform distribution over the current batch for efficient training or other biased distributions for hard negatives. And $\tau$ is the temperature parameter that controls the strength of penalties we enforced on hard negative samples
and $\delta(\cdot, \cdot)$ measures the similarity of two target views with cosine function.

\subsubsection{Improving User Representation via Information Bottleneck Contrastive Learning}\label{subsec:model_opt_u}

Similar as Section \ref{subsec:model_opt_e}, to equip the pre-trained recommender with cluster-level semantic \wrt user representation and make it more robust to inevitable noises in real-world environments, we propose to augment the user-side representation with intent-enhanced view.
Specifically, we adopt a $d$-dimensional learnable matrix $\mathbf{I} \in \mathbb{R}^{m \times d}$ to denote the latent intents for users, where $m$ is the number of intents. 
For each user $u$, we calculate the normalized similarity vector \wrt the $m$ intents as follows:
\begin{equation}
    \bm{s}_u = \text{SoftMax}(\mathbf{I} \cdot \bm{h}_u).
    \label{eq:sim}
\end{equation}
Then we obtain its intent-level view as follows:
\begin{equation}
\label{eq:pt_view}
    \hat{\bm{h}}_u = \mathbf{I}^T \cdot \bm{s}_u.
\end{equation}

However, unlike items that are static in the recommendation process, users act dynamically \wrt different recommended items hence we need to retain information of user representations that is most useful for a recommendation task. Inspired by Information Bottleneck that stems from information theory that states if we discard information from the original input which is not useful for a given task, it will enhance the robustness of the derived representation for downstream tasks. Therefore, different from traditional contrastive learning that encourages the commonality between representations of different views, instead we propose to encourage the divergence of the original view and the augmentation view while ensure that the augmentation view retains the maximal information that is relevant to our recommendation task. By replacing the $\bm{h}_{u}$ in Eq. \ref{eq:pretrain_yue} with $\hat{\bm{h}}_u$ and combining the revised $\mathcal{L}_{\mathcal{ET}}$, the Information Bottleneck Contrastive Learning can be defined as follows:
\begin{equation}
 \label{eq:loss_ib}
    \begin{split}
    \mathcal{L}_{IB} &= \mathcal{L}_{\mathcal{ET}} + \lambda_1 \sum_{u \in \mathcal{U}} \left(  \log{\frac{\exp{(\delta(\bm{h}_u, \hat{\bm{h}}_u)/\tau)}}{\sum_{u' \sim \mathcal{P}_{neg}}\exp{(\delta(\bm{h}_u, \hat{\bm{h}}_{u'})/\tau)}}} \right. \\
    & \left. + \log{\frac{\exp{(\delta(\hat{\bm{h}}_u, \bm{h}_u)/\tau)}}{\sum_{u' \sim \mathcal{P}_{neg}}\exp{(\delta( \hat{\bm{h}}_u, \bm{h}_{u'})/\tau)}}}  \right). \\
    \end{split}
\end{equation}

\subsubsection{Improving Decision Path via Hierarchical Structure}\label{subsec:model_opt_hier}
With the user / item representations optimized via exemplar-level
and information bottleneck based contrastive learning, we formally define the
decision path network from user intents to item exemplars and optimize the decision path selection via sparsity regularization and in the meantime takes
the node representations at both sides of the decision paths into consideration.

Specifically, consider there are amounts of paths among user intents and item exemplars, to explicitly model the path selection process we utilize a mask matrix $\mathbf{Z} \in \mathbb{R}^{n \times m}$ of binary variables where $\mathbf{Z}_{ij} = 1$ suggests that the edge between intent $i$ and exemplar $j$ is reserved while 0 otherwise. In practice, one core problem here is how to update each binary element $\mathbf{Z}_{ij}$. Inspired by \cite{l02018louizos}, we propose to model $\mathbf{Z}_{ij}$ as latent random variables from parameterized distributions and optimize it with our model parameters together. 
Let $u$ be a random variable sampled from a uniform distribution,  $\varepsilon$ be a binary concrete
random variable distributed in (0, 1) and $log\alpha$, $\beta$ be location and temperature parameters respectively, after stretching it into interval $(\gamma,\zeta)$\footnote{$\beta$, $\gamma$ and $\zeta$ is usually set to 2/3, -0.1, 1.1 respectively.} and transforming it via the hard sigmoid function, the derived distribution can keep exact zeros in its parameters and enable gradient optimization, with which we can come up with the $L_0$ regularization to explicitly model and keep the most important decision paths among user intents and item exemplars.
\begin{equation}
    \begin{split}
        \varepsilon =\sigma ((\log u- \log(1-u)+ \log\alpha )/\beta ), \ \ \  u\sim U(0,1) \\
        \bar{\varepsilon }=\varepsilon (\zeta -\gamma )+\gamma\\
        Z=g(\bar{\varepsilon }) = \min(1,\max(0,\bar{\varepsilon })).
    \end{split}
\end{equation}

\subsection{Learning of {\model}} \label{subsec:model_loss}

\paratitle{Pre-training Stage in Source Domains}
By integrating GNN-based pre-training strategy with hierarchical decision path enhanced representation learning module, the overall objective can be defined as follows:
\begin{equation}
    \label{eq:loss_source}
    \mathcal{L}_{\mathcal{PT}} = \mathcal{L}_{\mathcal{IB}} + \lambda_2 \cdot \mathcal{L}_{\mathcal{ECL}} + \lambda_3 \cdot \left \| Z \right \|_{0},
\end{equation}
where $\lambda_2$ and $\lambda_3$ are positive numbers denoting the relative
weight of each term.

\paratitle{Fine-tuning Stage in Target Domains} 
Formally, with a target domain $\mathcal{T} \in \mathbb{T}$, we easily initialize a recommender by utilizing the pre-training parameters, which is capable of producing representations of users and items. 
And then, given a user-item pair $\langle u, i\rangle$ in target domain $\mathcal{T}$, the user representation could be naturally obtained through universal encoding (denoted as $\hat{\bm{h}}_u^\mathcal{T}$), while we obtain the final representation for item $i$ (denoted as $\bm{h}^\mathcal{T}_i$) by adopting a mean pooling operation~\footnote{The straightforward mean pooling operation helps avoid amounts of delays for online inference, which could be naturally extended to complicated attentive aggregation.} on the representations of entities associated with item $i$.

Thus, the preference score is given by MLP and DeepFM~\cite{deepfm} as follows:
\begin{equation} \label{eq:y_target}
    \hat{y}_{ui} = \sigma(\text{MLP}(\hat{\bm{h}}_u^\mathcal{T} || \bm{h}^\mathcal{T}_i||\text{DeepFM}(\bm{x}^\mathcal{T}_u||\bm{x}^\mathcal{T}_i))),
\end{equation}
where $\bm{x}^\mathcal{T}_u$ and $\bm{x}^\mathcal{T}_i$ denote the profile information of user $u$ and item $i$ in target domain $\mathcal{T}$.  
Similarly, we adopt a  cross entropy loss as the final objective:
\begin{equation}  \label{eq:loss_target}
    \mathcal{L}_{\mathcal{FT}} = \sum_{\langle u, i, y_{ui}\rangle \in \mathcal{H}^{\mathcal{T}}} \mathcal{C}(y_{ui}, \hat{y}_{ui}).
\end{equation}
In general, the fine-tuning stage can be 
extended to consider the CVR estimation
by adopting the ESSM-like architecture~\cite{esmm}. 
Our approach also can effortlessly adapt to strictly cold-start scenarios that are common in online service platforms, by directly calculating the inner product of $\hat{\bm{h}}_u^\mathcal{T}$ and $\bm{h}^\mathcal{T}_i$.

\section{Experiments}
\vspace{-2em}
\begin{table}[H]\centering
\setlength{\tabcolsep}{0.5mm}
{

\caption{Overall performance evaluation on four target domains. The best results are highlighted in boldface. } 
\label{tab:overall_result2}
\begin{tabular}{c|c|c|c|c|c|c|c|c|c}

\toprule
\multirow{2}{*}{Tasks} & \multirow{2}{*}{Methods}
 & \multicolumn{2}{c|}{Exercise} 
& \multicolumn{2}{c|}{Leasing} 
& \multicolumn{2}{c|}{E-governance} 
& \multicolumn{2}{c}{E-mall} 
\\ 
\cmidrule{3-10}
 {} & {} 
 & {Hit} & {NDCG} 
 & {Hit} & {NDCG} 
 & {Hit} & {NDCG} 
 & {Hit} & {NDCG} 
 \\
\midrule	
&{DeepFM~\cite{deepfm}} 
& {0.2402} & {0.0406} 
& {0.2572} & {0.0480} 
& {0.1026} & {0.0136}
& {0.2195} & {0.0399} 
\\
\cmidrule{2-10} 
{} &{GRU4Rec~\cite{gru4rec2016}} 
& {0.2376} & {0.0417} 
& {0.2428} & {0.0437} 
& {0.0936} & {0.0125}
& {0.2468} & {0.0449} 
\\
\cmidrule{2-10}
{} &{BERT4Rec~\cite{bert4rec2019}} 
& {0.2311} & {0.0396} 
& {0.1348} & {0.0244} 
& {0.1077} & {0.0153}
& {0.2303} & {0.0411} 
\\
\cmidrule{2-10}
{} &{Comirec~\cite{comirec2020}} 
& {0.2486} & {0.0426} 
& {0.1058} & {0.0207} 
& {0.1872} & {0.0273}
& {0.2163} & {0.0387} 
\\
\cmidrule{2-10}
{} &{SASRec~\cite{sasrec2018}} 
& {0.3144} & {0.0546} 
& {0.3029} & {0.0534} 
& {0.1538} & {0.0235}
& {0.3142} & {0.0524} 
\\
\cmidrule{2-10}
{Few-shot} &{PeterRec~\cite{peterrec2020}} 
& {0.2501} & {0.0440} 
& {0.2746} & {0.0500} 
& {0.2590} & {0.0382}
& {0.2735} & {0.0498} 
\\
\cmidrule{2-10}
{Rec.} &{PAUP~\cite{paup2022}} 
& {0.2996} & {0.0516} 
& {0.2486} & {0.0455} 
& {0.1154} & {0.0170}
& {0.2506} & {0.0459} 
\\
\cmidrule{2-10}	
{} &{LightGCN~\cite{lightgcn2020}} 
& {0.6221} & {0.1025} 
& {0.3609} & {0.0589} 
& {0.2526} & {0.0370}
& {0.4059} & {0.0684} 
\\
\cmidrule{2-10}
{} &{DisenGCN~\cite{disengnn2022}} 
& {0.5962} & {0.0869} 
& {0.4268} & {0.0716} 
& {0.2167} & {0.0313}
& {0.3359} & {0.0544} 
\\
\cmidrule{2-10}
{} &{SimGCL~\cite{simgcl2022}}
& {0.5057} & {0.0962} 
& {0.3167} & {0.0551} 
& {0.2244} & {0.0326}
& {0.4046} & {0.0676} 
\\
\cmidrule{2-10}
{} &{M2GNN~\cite{m2gnn2023}} 
& {0.4818} & {0.0717} 
& {0.4594} & {0.0752} 
& {0.2846} & {0.0438}
& {0.3346} & {0.0558} 
\\
\cmidrule{2-10}
{} &{HIER} 
& {\textbf{0.6472}} & {\textbf{0.1062}} 
& {\textbf{0.5239}} & {\textbf{0.0848}} 
& {\textbf{0.4038}} & {\textbf{0.0615}}
& {\textbf{0.4377}} & {\textbf{0.0726}} 
\\
\cmidrule{1-10}
{Zero-shot} & {Tiger~\cite{tiger2022}} 
& {0.1563} & {0.0374} 
& {0.3022} & {\textbf{0.0734}}  
& {0.2141} & {0.0474} 
& {0.2354} & {0.0484} 
\\
\cmidrule{2-10}
{Rec.} &{HIER} 
& {\textbf{0.2934}} & {\textbf{0.0549}} 
& {\textbf{0.3109}} & {0.0626}  
& {\textbf{0.2833}} & {\textbf{0.0598}} 
& {\textbf{0.3626}} & {\textbf{0.0849}} 
\\

\bottomrule
\end{tabular}}
\end{table}
\vspace{-1em}

\subsubsection{Datasets}
Considering that the proposed {\model} mainly focuses on the industrial cross-domain recommendation issue in online service platforms, we evaluate it by employing real-world industrial datasets collected from different domains in Alipay. 
Specifically, one month's data from \textbf{Payment} ($\sim$500 million users and $\sim$5.6 million entities with $\sim$6000 million interactions), 
\textbf{Search} ($\sim$30 million users and $\sim$150 thousands entities with $\sim$50 million interactions) and click behaviors on the 
\textbf{Homepage Feed} ($\sim$200 million users and $\sim$22 thousands entities with $\sim$1000 million interactions) are collected as the source domains.
For the target domains, due to data sparsity, we collect two month's data from the following four \textit{scenarios with manifest intents}: \textbf{Exercise} service ($\sim$17 million interactions between $\sim$1 million
users and $\sim$ 300 items), 
\textbf{Leasing} service ($\sim$1.4 million interactions between $\sim$140 thousand
users and $\sim$ 2500 items), 
\textbf{E-governance} service ($\sim$1 million interactions between $\sim$100 thousand
users and $\sim$ 12 thousand items), 
and \textbf{E-mall} service ($\sim$14 million interactions between $\sim$140 thousand
users and $\sim$ 120 thousand items).
Moreover, to bridge up the domain gap, a knowledge graph involving more than ten millions of entities and more than one hundred millions of triplets \cite{kddyiru} is incorporated for cross-domain recommendation, which adopts a 32-dimensional embedding from a pre-trained BERT as the original features for each entity. As for target domains in few-shot setting, several attributes are also extracted for model training, \ie user-side attributes include age, occupation, behaviors in current domains and so on while item-side attributes contain item id, category, brand and so on.




\subsubsection{Overall Performance}
From the empirical results in few-shot and zero-shot (Table~\ref{tab:overall_result2}) transferable recommendation tasks, the major findings can be summarized as follows: 
i) Significant performance gains of {\model} on four target domains with manifest intents in both few-shot and zero-shot settings can be found, owing to that {\model} explicitly model user’s hierarchical decision path and improve the representation of users and items along this path. 
The overall improvements \wrt Tiger in zero-shot recommendation further demonstrate that {\model} could greatly warm up a newly-launched scenario without fine-tuning. 
ii) Owing to the tremendous differences \wrt intents, frequencies of target domains, generally we find that graph learning approaches perform better when compared to sequential learning based competitors especially in low-frequency intents such as E-governance, which demonstrate the generality of higher-order structural information in knowledge graph, as a useful auxiliary information, helps to abstract transferable knowledge from source domains to empower various downstream target domains with manifest intents.

\subsubsection{Ablation study} 
\begin{figure}[t]
\vspace{-0.5em}
	\centering
	\subfigure[NDCG@25]{\includegraphics[width=0.42\columnwidth]{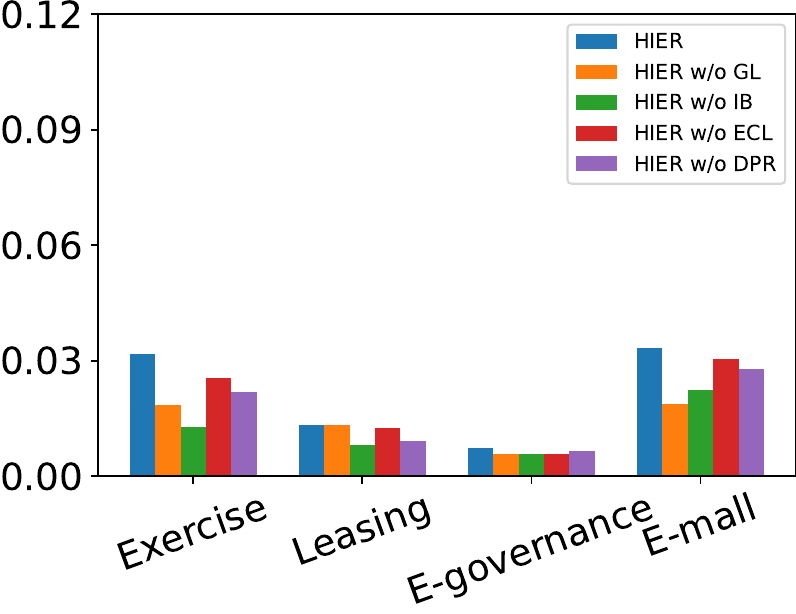}}
	\subfigure[NDCG@50]{\includegraphics[width=0.42\columnwidth]{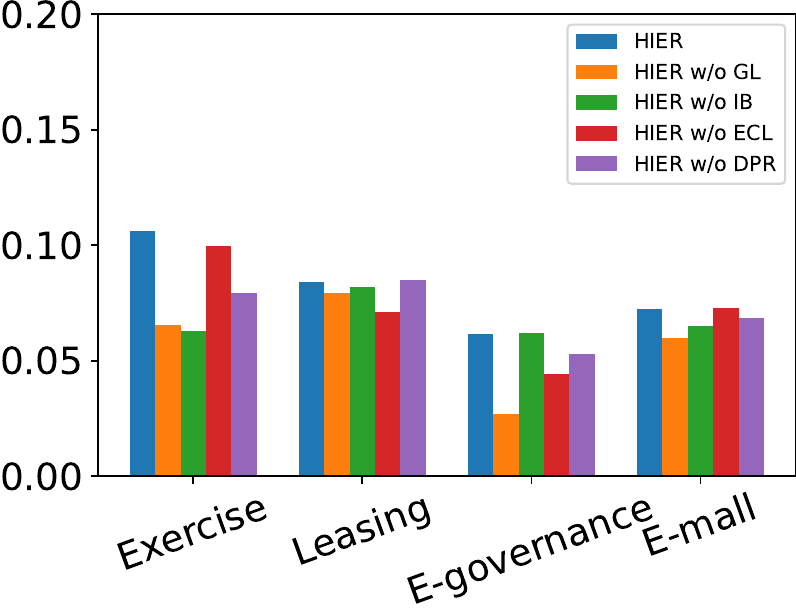}}
  \caption{Ablation study}
	\label{fig:ab_study}
	\vspace{-0.5em}
\end{figure}

We examine the effectiveness of each component in {\model} by preparing the following variants: 
i) \textbf{{\model} w/o GL} (removing the knowledge graph enhanced Graph Learning),
ii) \textbf{{\model} w/o IB} (removing the Information Bottleneck Contrastive Learning),
iii) \textbf{{\model} w/o ECL} (removing the Exemplar-enhanced Contrastive Learning),
iv) \textbf{{\model} w/o DPR}  (removing the Decision Path Regularization that the decision path selection process downgrades to fully-connected network).
We present the comparison results in Figure \ref{fig:ab_study}. Generally, the performance would drop a lot when one component is discarded, which justify the integration of components in {\model}. Specifically, we notice that in many cases {\model} w/o GL performs worst, which further certify that the knowledge graph, as a cornerstone for representation learning enhancement, empowers the subsequent components.

\subsubsection{Analysis of Exemplar-level Contrastive Learning}
\begin{figure}[t]
	\centering
	\subfigure[Exercise]{\includegraphics[width=0.24\columnwidth]{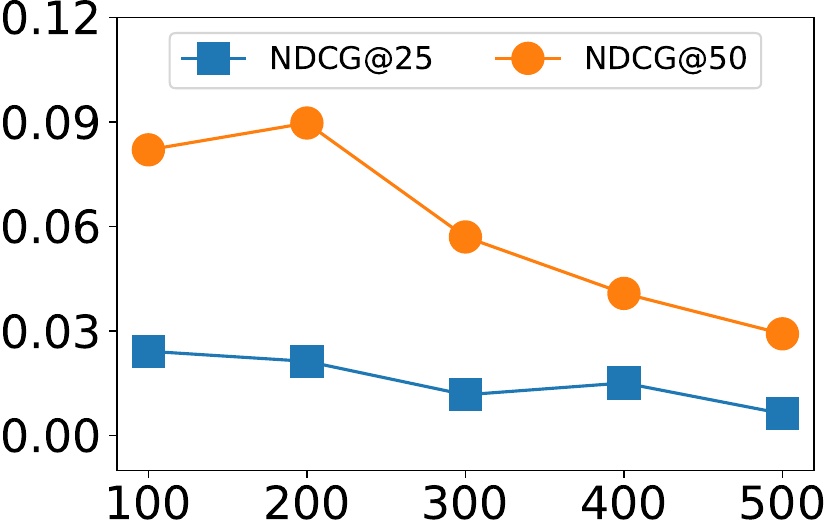}}
	\subfigure[Leasing]{\includegraphics[width=0.24\columnwidth]{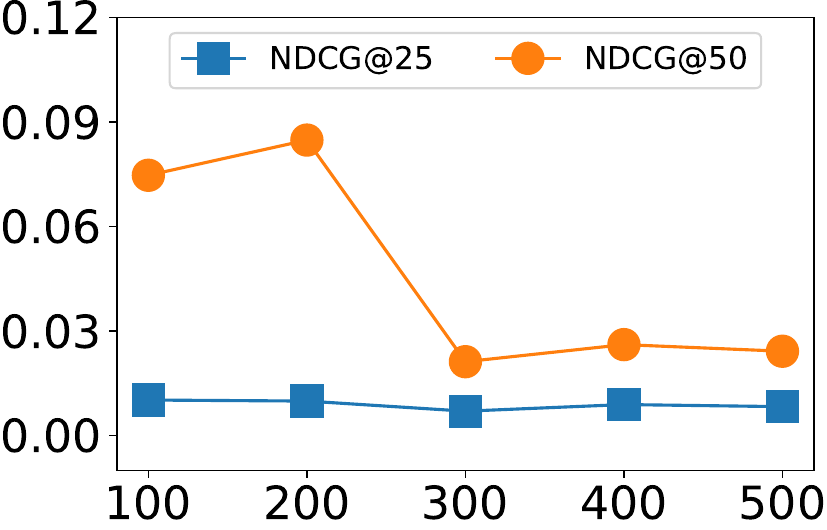}}
        \subfigure[E-governance]{\includegraphics[width=0.24\columnwidth]{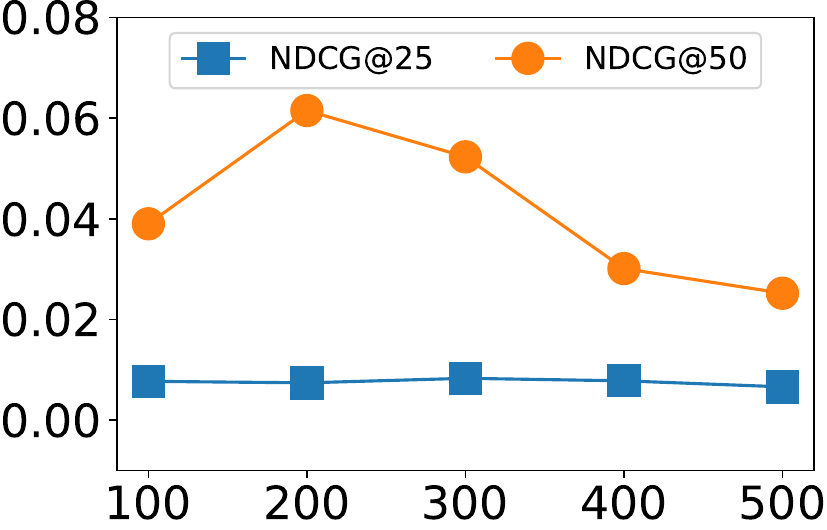}}
        \subfigure[E-mall]{\includegraphics[width=0.24\columnwidth]{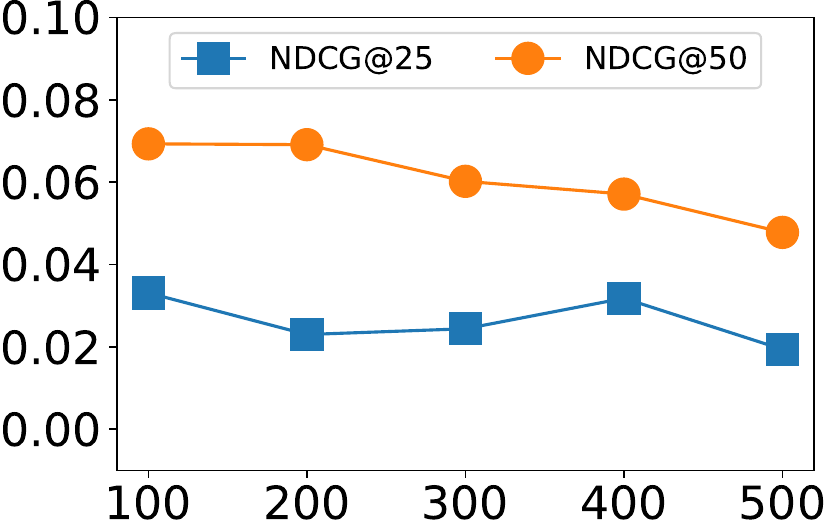}}
  \caption{ The influence of exemplar number \wrt NDCG.}
	\label{fig:exemplar}
	\vspace{-1.75em}
\end{figure}

Here, we investigate into the impact of exemplar number by varying it from 100 to 500 and present the experimental result in Figure \ref{fig:exemplar}. We notice that {\model} always keeps best performance with a middle number of exemplars, while too large value of exemplar number hurt the performance due to inadequate small granularity of clustering space for items and too small value of exemplar number influence the performance owing to the inappropriate semantics a large exemplar carries.

\vspace{-0.5em}
\subsubsection{Analysis of Information Bottleneck Contrastive
Learning}
\begin{figure}[h]
\vspace{-1.25em}
	\centering
	\subfigure[Exercise]{\includegraphics[width=0.24\columnwidth]{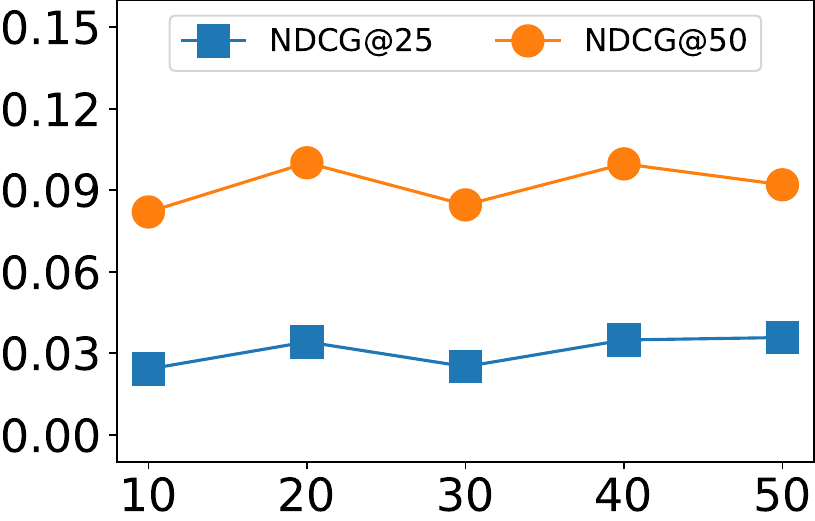}}
	\subfigure[Leasing]{\includegraphics[width=0.24\columnwidth]{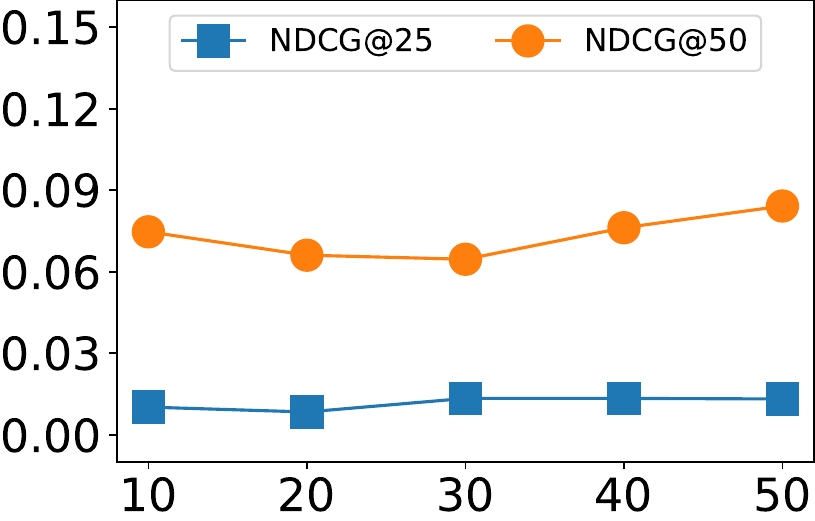}}  
        \subfigure[E-governance]{\includegraphics[width=0.24\columnwidth]{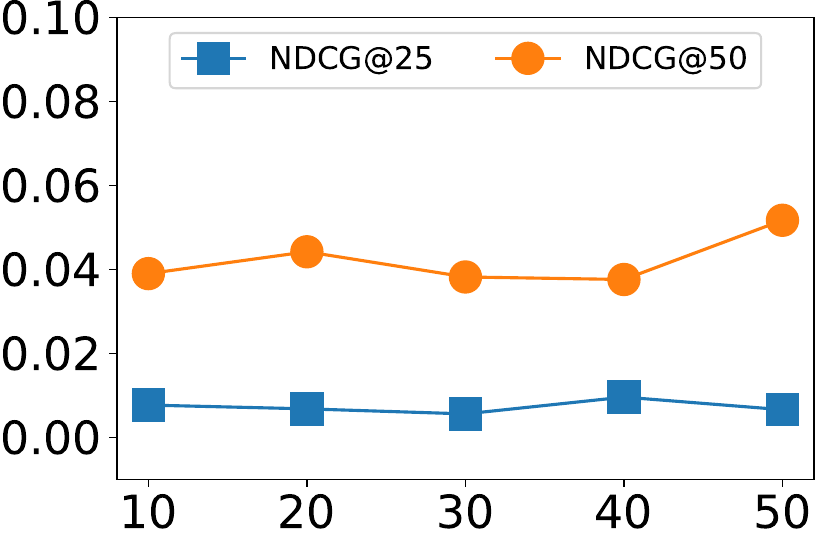}}
        \subfigure[E-mall]{\includegraphics[width=0.24\columnwidth]{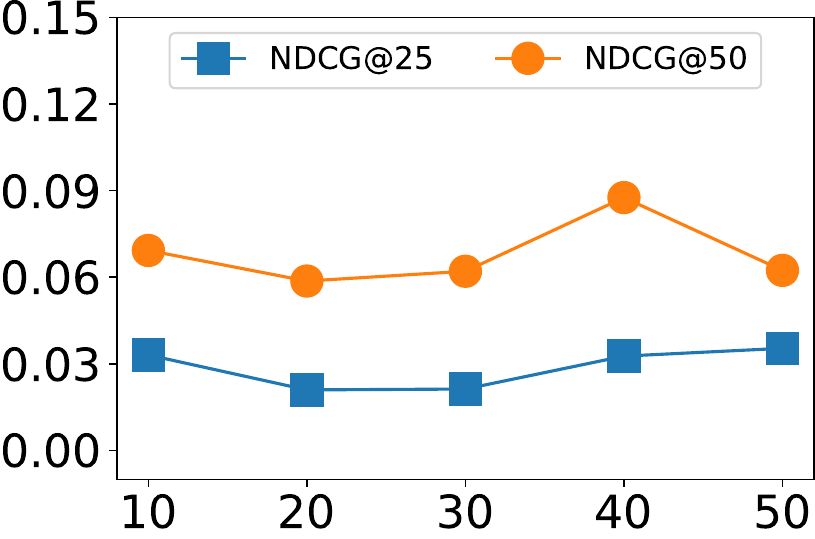}}
  \caption{ The influence of intent number \wrt NDCG.}
	\label{fig:intents}
	\vspace{-1.75em}
\end{figure}

To investigate the impact of intent number, we vary it from 10 to 50 and present the experimental result in Figure \ref{fig:intents}. It is noted that {\model} always best performance with a larger number of intents due to reason that when human beings make decisions the underlying intents are rather complicated. Taking riding a bicycle as an example, when a person interacts with a high-end bicycle the interests toward riding may be the dominant intent while when a person interacts with a sharing-bike the commuting needs may be the dominant one. However, with a small number of intents, it is hard to capture the subtle difference between them thus downgrading the model performance.

\subsubsection{Online Performance}
\begin{figure}[H]
	\centering
	\subfigure[Exercise]{\includegraphics[width=0.24\columnwidth]{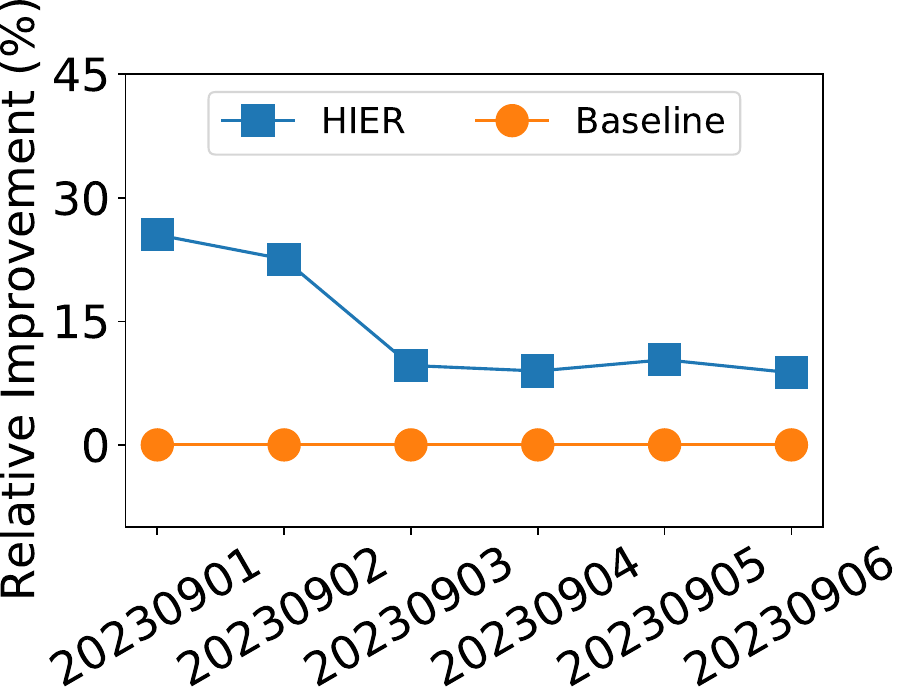}}
	\subfigure[Leasing]{\includegraphics[width=0.24\columnwidth]{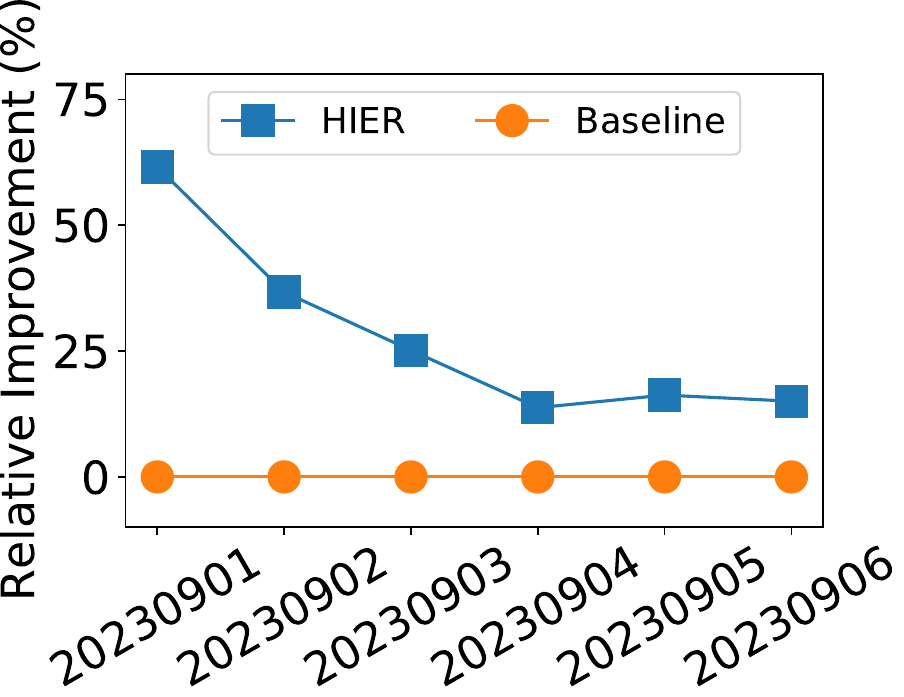}}     
    \subfigure[E-governance]{\includegraphics[width=0.24\columnwidth]{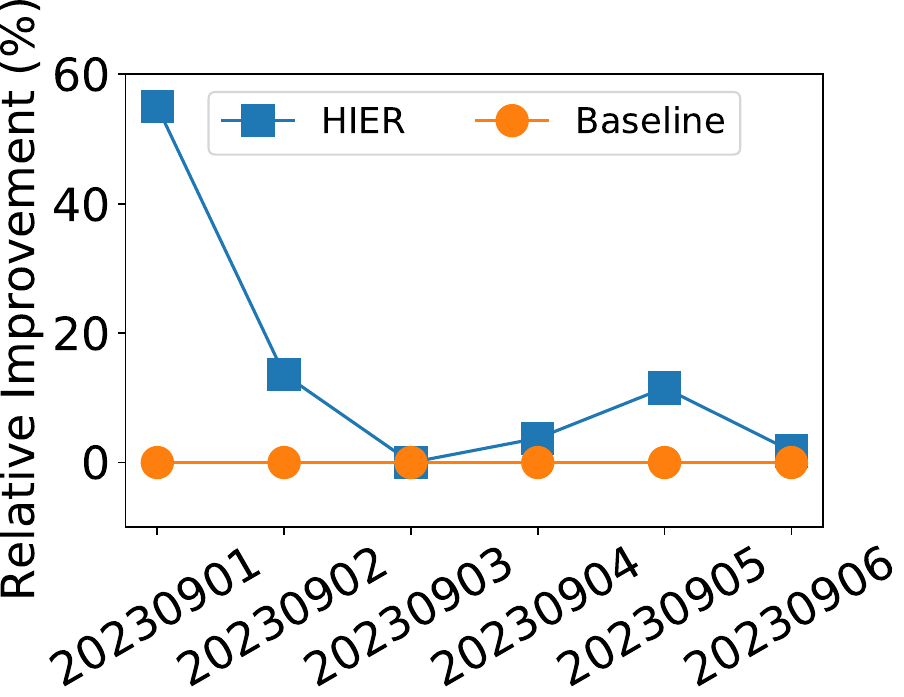}}
     \subfigure[E-mall]{\includegraphics[width=0.24\columnwidth]{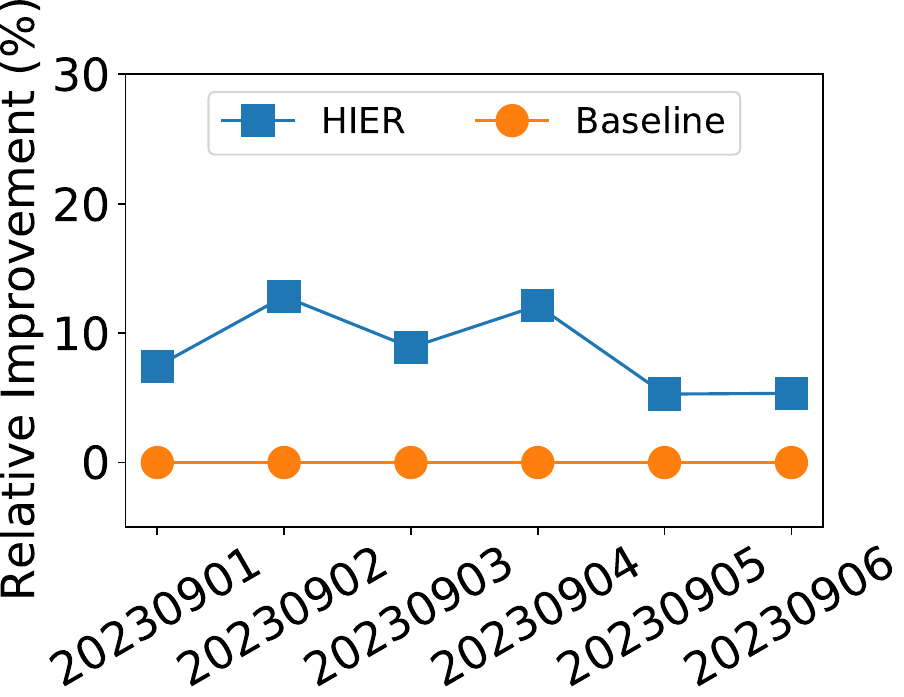}}
  
  \caption{Online performance on real-world scenarios.}
	\label{fig:online_ab}
	\vspace{-1.5em}
\end{figure}

We evaluate {\model} via real traffic in four downstream scenarios by competing against the deployed DeepFM-based baseline. We perform online evaluation from ``2023/09/01'' to `2023/09/06' with the widely-adopted CTR 
as the core metric, and report the experimental results in  Figure \ref{fig:online_ab}.
We find that  {\model} achieves significant
performance gains by 11.63\%, 22.90\%, 5.68\%, and 8.26\% 
with a significance level of 95\% in the four scenarios, which further demonstrates the effectiveness of {\model}  in real-world service platforms.

\vspace{-1.5em}

\section{Conclusion}
In this paper, we propose {\model} that takes user's 
hierarchical decision path into consideration when pre-training industrial recommenders via multi-source behaviors.
Extensive experiments on real-world industrial datasets illustrate that {\model} significantly outperforms state-of-the-art baselines in various scenarios.

%
%
%
%

\end{document}